%% file: main.tex
\documentclass[10pt,twocolumn,letterpaper]{article}

\usepackage{iccv}              

\definecolor{iccvblue}{rgb}{0.21,0.49,0.74}
\usepackage[pagebackref,breaklinks,colorlinks,allcolors=iccvblue]{hyperref}
\usepackage{url}
\usepackage{booktabs} 
\usepackage{multirow}
\usepackage{graphicx}
\usepackage{wrapfig}
\usepackage{array}
\usepackage{makecell}
\usepackage[accsupp]{axessibility}  
\usepackage{stfloats}
\def\paperID{14959} 
\def\confName{ICCV}
\def\confYear{2025}

\title{Multimodal LLMs as Customized Reward Models for Text-to-Image Generation}

\author{Shijie Zhou$^1$\thanks{Work done at University at Buffalo through University Collaborations.},~~Ruiyi Zhang$^2$\thanks{Project Lead and Corresponding Author.},~~~Huaisheng Zhu$^3$,~~ Branislav Kveton$^2$,\\ Yufan Zhou$^4$\thanks{Work done during Adobe Research.}, ~~Jiuxiang Gu$^2$, ~~Jian Chen$^1$, ~~Changyou Chen$^1$ \\
    $^1$University at Buffalo,~~~~$^2$Adobe Research,~~~~$^3$Pennsylvania State University~~~~$^4$Luma AI \\
    {\tt\small \{ryzhang.cs\}@gmail.com}, {\tt\small \{shijiezh, changyou\}@buffalo.edu}
}
\newcommand{\method}{LLaVA-Reward}

\input{math}

\begin{document}
\maketitle
\input{sec/0_abstract}    
\input{sec/1_intro}
\input{sec/2_related}

\begin{figure*}
    \centering
    \includegraphics[width=0.85\linewidth]{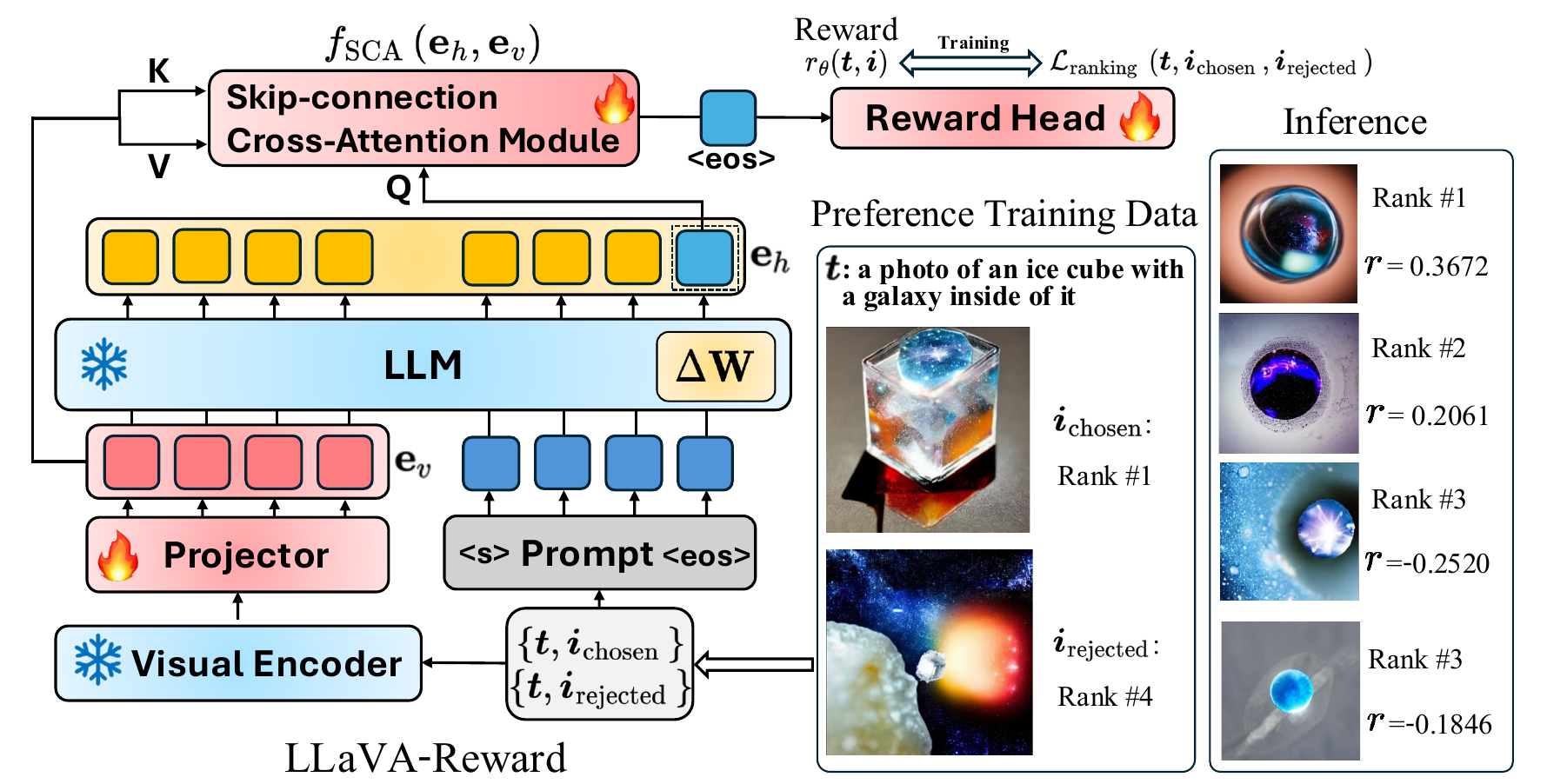}
    \vspace{-0.25cm}
    \caption{Model overview of \method{}. \method{} is fine-tuned on the lightweight Phi-3.5-vision using the pairwise preference data without lengthy instructions via preference ranking loss, \textit{e.g.}, Bradley-Terry loss. The reward $r_\theta(\boldsymbol{t}, \boldsymbol{i})$ is output from the proposed Skip-connection Cross-Attention (SkipCA) module, which connects projected visual token $\mathbf{e}_v$ and the final hidden state $\mathbf{e}_h$. The preference/reward outputs of \method{} can be used for downstream tasks, such as text-to-image generation and evaluation.}
    \vspace{-0.5cm}
    \label{fig:model}
\end{figure*}
\input{sec/3_method}
\input{sec/4_experiment}

\input{sec/5_conclusion}

{   
    \small
    \bibliographystyle{ieeenat_fullname}
    \bibliography{main.bbl}
}

\input{sec/X_suppl}

\end{document}

%% file: math.tex
\usepackage{amsmath,amsfonts,bm}

\def\eqref#1{equation~\ref{#1}}

\def\1{\bm{1}}

\def\vi{{\bm{i}}}

\def\vt{{\bm{t}}}

\DeclareMathAlphabet{\mathsfit}{\encodingdefault}{\sfdefault}{m}{sl}
\SetMathAlphabet{\mathsfit}{bold}{\encodingdefault}{\sfdefault}{bx}{n}



%% file: sec/0_abstract.tex
\begin{abstract}
We introduce \textbf{\method{}}\footnote{Project page: \href{https://github.com/sjz5202/LLaVA-Reward}{https://github.com/sjz5202/LLaVAReward}.}
, an efficient reward model designed to automatically evaluate text-to-image (T2I) generations across multiple perspectives, leveraging pretrained multimodal large language models (MLLMs). Existing MLLM-based approaches require instruction-following data for supervised fine-tuning and evaluate generation quality on analyzing text response, which is time-consuming and difficult to train. To address this problem, we propose \method{}, which directly utilizes the hidden states of MLLMs given text-image pairs. 
To enhance the bidirectional interaction between visual and textual representations in decoder-only MLLMs, we further propose adding a Skip-connection Cross Attention (SkipCA) module. This design enhances text-image correlation reasoning by connecting early-layer visual features with later-layer hidden representations.
In addition, \method{} supports different types of preference data for efficient fine-tuning, including paired preference data and unpaired data. 
We train \method{} on four evaluation perspectives: text-image alignment, fidelity/artifact, safety, and overall ranking.  
Empirical results demonstrate that \method{} outperforms conventional and MLLM-based methods in generating human-aligned scores for automatic evaluations and inference-time scaling in text-to-image generations.
\end{abstract}

%% file: sec/1_intro.tex
\section{Introduction}
\label{sec:intro}
Reward modeling and preference learning~\citep{zhang2024general,liu2024skywork,wang2024helpsteer2} have been crucial components in enhancing the alignment between foundation models and human preferences~\citep{ouyang2022training,fan2023dpok,schulman2017proximal,meng2024simpo,black2023training}. 
By learning from human preference and annotated scores, reward models can guide the training process of foundation models by encouraging generations that better align with human preference, while simultaneously penalizing outputs that deviate from desired behaviors. Recent works have widely explored fine-tuning LLMs as reward models within reinforcement learning from human feedback (RLHF) frameworks~\citep{schulman2017proximal,ouyang2022training} or preference optimization~\citep{rafailov2024direct}. 

Text-to-image generation has evolved significantly since the advent of GANs~\citep{goodfellow2020generative}, with notable advancements from models such as DALL·E~\citep{ramesh2021zero}, LAFITE~\citep{zhou2022towards}, and CogView~\citep{ding2021cogview}. More recently, diffusion models~\citep{dhariwal2021diffusion, ho2020denoising, sohl2015deep}, including Stable Diffusion~\citep{rombach2022high}, have achieved state-of-the-art results. 
Despite these explorations, the development of effective reward models for text-to-image generation still faces significant challenges and limitations. Early CLIP-based reward models, such as CLIPScore, PickScore-v1~\citep{kirstain2023pick}, HPSv2~\citep{wu2023human}, and ImageReward~\cite{xu2024imagereward}, fine-tune the CLIP model with similarities in the text-image pair. However, due to CLIP's tendency to behave like a bag-of-words model~\citep{yuksekgonul2023when} and its limited generalization capabilities, these models struggle with complex image-text relationships. 
In the context of the reward model for text-to-image generation~\citep{rombach2022high}, recent research focuses on CLIP or BLIP-based methods~\citep{hessel2021clipscore,xu2024imagereward}, as well as visual question-answering (VQA) based scoring~\citep{tan2024evalalign}, which can only give assessments on specific perspectives, such as text-image alignment or safety. 
Recent MLLM-based reward models rely on carefully designed system prompts to induce reward in the generated answer~\citep{li2025t2isafety,helff2024llavaguard,tan2024evalalign} or in the likelihood of reward token \textit{, e.g.}, ``good", ``bad"~\citep{wu2023q,li2024removing,lin2024evaluating}. However, lengthy system prompts with complex requirements are required, making it inefficient during training and inference. Another concern is that both VQA-based and token-based methods need a specifically designed label, \textit{i.e.}, a decisive or rating word, making them difficult to exploit relative preference data.

To address these limitations, we propose \method{}, an efficient and flexible multimodal reward model designed to evaluate text-to-image generation from diverse perspectives. Unlike prior MLLM-based approaches that rely heavily on instruction-following data~\cite{tan2024evalalign} or VQA-style next-token likelihood~\cite{lin2024evaluating}, \method{} directly leverages the hidden states of pre-trained MLLMs with the text-image pairs as inputs to produce reward/evaluations without the need for lengthy rating instructions. 
\method{} built on pre-trained MLLM with LoRA~\cite{hu2021lora} adapters using human preference data through Bradley-Terry rankings~\cite{bradley1952rank}.
In addition, \method{} enhances the cross-modal reasoning by adding the Skip-connection Cross-attention module (SkipCA), which can integrate visual features in the early stages with textual representations in the later stages.
In summary, our contributions are as follows,
\begin{itemize}
     \item We propose \method{}, a novel multimodal reward model, built on pre-trained MLLM with LoRA adapters, for efficient fine-tuning of human preference data without requiring complex system prompts. 
     \item \method{} is the first model that can perform an efficient auto-evaluation on multiple perspectives, including alignment, fidelity / artifact, aesthetics, and safety.  
    \item \method{} achieves state-of-the-art performance in all public benchmarks for text-to-image evaluation.  
    \item We also demonstrate that using \method{} within inference-time scaling~\citep{ma2025inference,singhal2025general} can significantly improve image quality, surpassing existing reward models, such as HPSv2 and ImageReward.
\end{itemize}

%% file: sec/2_related.tex
\section{Related Works}
\label{sec:related_work}
\paragraph{Automatic Text-to-image Evaluation}
Perceptual metrics such as IS~\citep{barratt2018note} and FID~\citep{heusel2017gans} primarily assess image coherence, recent research has shifted towards developing multimodal judges for text-to-image generation, where diverse aspects between the images and reference texts exist for evaluations, including text-image alignment, fidelity, aesthetics, and safety, etc. Early practice such as CLIPScore~\citep{hessel2021clipscore}, Pick-a-Pic~\citep{kirstain2023pick}, HPS~\citep{wu2023human}, ImageReward~\citep{xu2024imagereward} directly employ CLIP~\citep{radford2021learning} or incorporate human preference rating into CLIP fine-tuning for better text-to-image assessment. However, these works directly fine-tune CLIP models, which tend to behave like bag-of-words models~\cite {yuksekgonul2023when} and exhibit limitations in generalization. To address the limitations of conventional CLIP-based models, recent efforts have leveraged multimodal large language models (MLLMs) for assessing text-to-image generation, focusing on text-image alignment and safety.  
\vspace{-1em}
\paragraph{Evaluation based on visual question answering (VQA)} 
VQA-based text-to-image evaluation approaches~\citep{ku2023viescore,tu2024automatic,tan2024evalalign,helff2024llavaguard,li2025t2isafety,qu2024unsafebench,ghosh2023geneval} employ detailed evaluation instructions tailored to specific perspectives to allow MLLM to produce evaluations for the input text-image pair that adhere to the intended criteria in the prompts. Specifically, 
VIEScore~\citep{ku2023viescore} split text-to-image consistency and quality evaluation into multiple sub-scores in its designed prompts and leverage GPT-4o for synthesized image analysis. 
T2I-Eval~\citep{tu2024automatic} develops more fine-grained subtasks in their evaluation prompts and distills GPT-4o's evaluations toward these subtasks into the open-source MLLM for better efficiency.
EvalAlign~\citep{tan2024evalalign} further enriches the rating levels with details and concrete rationales in the subtasks and collects the corresponding human feedback for fine-tuning. 
Another set of VQA-based text-to-image judges focuses on assessing the safety of visual content, such as LlavaGuard~\citep{helff2024llavaguard}, ImageGuard~\citep{li2025t2isafety} and PerspectiveVision~\citep{qu2024unsafebench}. Although all follow the prompt design of LlamaGuard~\citep{inan2023llama}, LlavaGuard enables the extra category and rationale generations behind the safety rating, ImageGuard enhances multimodal interactions by introducing cross-modal attention into MLLMs, PerspectiveVision finetunes MLLMs with its UnsafeBench for safety evaluation and classification. 
The detailed rule/policy prompt in VQA-based judges unavoidably brings additional computation costs, making it impractical for the RLHF training. Additionally, VQA-based MLLM judges produce less fine-grained and precise ratings due to the limitation of discrete scoring in training data.
\vspace{-1em}
\paragraph{Evaluation based on special token probability}
Token Probability-based text-to-image evaluation methods~\citep{wu2023q,li2024removing,wang2024mllm,lin2024evaluating} measure the generative likelihood of the decisional or rating token, such as ``yes", ``no", ``good", ``bad", responding to the evaluation prompt and taking the likelihood value to form the evaluation score. 
Q-ALIGN~\cite{wu2023q} models the human score as rating levels via cross-entropy loss and computes the expectation on different levels' predicting likelihood as the score for inference. 
VQAScore~\citep{lin2024evaluating} and LLaVAScore~\citep{li2024removing} both use the likelihood of the ``Yes" token in MLLM's output as the score for text image alignment. The difference is that VQAScore fine-tunes an encoder-decoder MLLM and LLaVAScore adapts the MLLM using visual compositional data with negative text-image pairs. 
CLUE~\citep{wang2024mllm} uses token probability-based scoring to form a chain of safety judges.

Compared with VQA-based methods, token-based methods are much more efficient. However, their training still relies on discrete assessment tokens using the cross-entropy loss, leading to biased evaluation scores. It is difficult to choose a golden token for all evaluation tasks. Furthermore, it is difficult for these methods to learn from samples with a small quality gap, where there are no tier labels.
In \method, we push the MLLM to learn how to rank text-image pairs using the Bradley-Terry ranking loss to produce the reward as the evaluation score, instead of the prediction of the desired assessment token. It enables for precise, more fine-grained, continuous assessments.

%% file: sec/3_method.tex
\section{LLaVA-Reward}
\label{sec:experiments}
\label{sec:method}
\method{} is an efficient reward model for the evaluation of text-to-image generation in terms of alignment, artifact, and safety. It leverages pre-trained multimodal large language models (MLLMs), taking solely the text-image pair as input, and utilizes the hidden state of MLLMs for the preference modeling. 
Given a target perspective $p \in \mathcal{P}$ for text-to-image evaluation and an image-text pair $\{\vi,\vt\}$, our goal is to compute a reward $r_{\theta_p}(\vi, \vt)$ or preference $\mathbb{P}_{\theta_p}(\vi_c \succ \vi_{r} \mid \vt)$, where text $\vt$ is the prompt or caption for the corresponding image $\vi$. The perspective set $\mathcal{P}$ includes diverse perspectives towards intrinsic properties of the image $\vi$, such as safety and fidelity, or cross-modal criteria, such as text-image alignment. 
\vspace{-1em}
\paragraph{Architecture Design} The architecture of \method{} is shown in \cref{fig:model}, where we select Phi-3.5-vision 4.2B~\citep{abdin2024phi} as the base MLLM. Please note that our method is general and any MLLM can be used as the base model. Phi-3.5-vision is chosen due to its lightweight and superior capacity for human-aligned vision-language understanding. We modify Phi-3.5-vision by incorporating an additional bidirectional reward reward head $f_r(\mathbf{e}_{h},\mathbf{e}_{v})$. This reward head takes the hidden layer (usually the final layer) embedding $\mathbf{e}_h$ of the end-of-sentence (EOS) token as inputs to output the reward. 

Some of existing methods~\cite{lin2024evaluating, tan2024evalalign, ku2023viescore} perform supervised fine-tuning on MLLM using instruction data, \textit{i.e.,} a triplet of a text-image pair, a system instruction, and a text answer with ground truth reward, which is usually time-consuming during inference time. \method{} chooses to abandon the text generation ability of the pre-trained MLLM and utilize the hidden embeddings (usually the last layer) to predict a reward $r$ from the designed reward head. It avoids lengthy and complex human-designed evaluation instructions. Hence, \method{} offers significantly better efficiency and flexibility.
\vspace{-1em}
\paragraph{Bidirectional Reward Head} In recent practices of LLM reward models for language~\citep{wang2024helpsteer2}, the reward head is typically implemented as a linear projection layer. However, due to the unidirectional causal attention mechanism in decoder-only MLLMs, visual tokens are not influenced by the later injected text tokens. It will undermine the reasoning capacity of the MLLM reward model on image-text correlation, especially in the perspective that the visual part should be emphasized in rewarding, such as image generation safety and fidelity. 

To enhance bidirectional awareness between visual and text tokens in decoder-only MLLMs, we replace the linear projection layer with a simple and effective Skip-connection Cross Attention Modal (SkipCA) module as the reward head, which is inspired by cross-modality attention (CMA)~\citep{li2025t2isafety}. 
Specifically, we denote the SkipCA module by $f_{\mathrm{SCA}}$, which is a standard Cross-Attention operator. It takes the projected visual tokens $\mathbf{e}_v$ from the visual projector as the key and value, and the hidden state of the deeper layer $\mathbf{e}_h$ as the query. $r_{\theta}(\vi, \vt)= f_r(\mathbf{e}_{h},\mathbf{e}_{v}) = g(f_{\mathrm{SCA}}(\mathbf{e}_h,\mathbf{e}_v))$ is predicted based on the output of the SkipCA module on the EOS token during inference, where $g$ is a linear layer projection to produce a scalar reward under Bradley-Terry model or a vector reward under General Preference model (GPM)~\citep{zhang2024general}. 
This skip layer modeling between $\mathbf{e}_h$ and $\mathbf{e}_v$ builds the bidirectional connections between visual and text tokens to benefit the text-image evaluation. 
Compared to CMA~\citep{li2025t2isafety}, which merges image tokens into text tokens of the same layer in multiple layers of the language model in MLLMs, skip-connection cross-attention in \method{} is a more direct and effective practice. As visual tokens contribute less in the deeper transformer layers of MLLMs~\citep{zhang2024treat}, the CMA of ImageGuard has a small impact in the deep layers. Thus, we extract visual tokens $\mathbf{e}_v$ right after the visual projector, which are much more visual-specific than deeper tokens, to cross-attention with final-layer text tokens $\mathbf{e}_h$ to increase visual awareness in the reward process. 
\vspace{-1em}
\paragraph{LoRA Adaptation for Different Perspectives}
We customize the MLLM-based reward model for each evaluation perspective $p$. \method{} utilizes the hidden embeddings (usually the last layer) to predict a reward $r$ and does not need perspective-specified instructions. To enable it to support multi-perspective evaluation, we adopt Phi-3.5-vision with LoRA adapters. Each adapter corresponds to a specific perspective, and we can easily change it for another perspective. Hence, \method{} shares most of the parameters of the pre-trained MLLMs, facilitating the efficiency in terms of time and memory.
\vspace{-1em}
\paragraph{\method{} Objective.}
\method{} can perform efficient and memory-friendly fine-tuning via Bradley-Terry ranking loss (for paired preference data) and classification loss (for unpaired affirmative data), such as UnsafeBench~\citep{qu2024unsafebench} for safety. For the task with paired text-to-image preference~\citep{chen2024mj} data $\{\vt,\vi_c,\vi_r\}$, containing a chosen image $\vi_c$ and a rejected image $\vi_r$ generated from the same prompt $\vt$ for perspective $p$, we finetune the pretrained MLLM $\theta_p$ with the standard pairwise ranking loss under the Bradley-Terry (BT) model with temperature $T$:
\begin{equation}
\begin{aligned}
\mathcal{L}_{\mathrm{rank}}
  &= \mathbb{E}_{( \vi_c, \vi_r, \vt)\sim\mathcal D_p}
     \Bigl[
        -\log \sigma\!\Bigl(
            \frac{s_{\theta_p}( \vi_c, \vt)
                  -s_{\theta_p}( \vi_r, \vt)}{T}
        \Bigr)
     \Bigr] ,
\end{aligned}
\label{eq:btloss}
\end{equation}
where $s_{\theta_p}(\vi, \vt)$ is the explicit reward from the MLLM parameterized by $\theta_p$, $\mathcal D_p$ is the text-to-image preference data on perspective $p$.
The form of $s_{\theta_p}(\vi, \vt)$ is flexible. In the case of the scalar reward output $r_{\theta_p}(\vi, \vt)$ of \method{}, we have $s_{\theta_p}\left(\vi, \vt\right)=r_{\theta_p}(\vi, \vt)$.
To better capture complex visual structures and text prompts, we consider preference embedding, following the General Preference Model (GPM)~\citep{zhang2024general}. This design enables an efficient and expressive representation of human preferences.
Specifically, with $m$-dimensional reward outputs $r_{\theta_p}(\vi_c, \vt)$ and $r_{\theta_p}(\vi_r, \vt) \in \mathbb{R}^m$, GPM forms the reward difference in \cref{eq:btloss} as: 
\begin{equation}
    s_{\theta_p}(\vi_c, \vt)-s_{\theta_p}(\vi_r, \vt)=\left\langle\mathbf{R}^{\succ} r_{\theta_p}(\vi_c, \vt), r_{\theta_p}(\vi_r, \vt)\right\rangle,
\end{equation}
where $\langle\cdot, \cdot\rangle$ denotes the inner product and $\mathbf{R}^{\succ}$ is the skew-symmetric preference operator, which for $m=2$ takes the form $\mathbf{R}^{\succ}=\left[\begin{array}{cc}
0 & -1 \\
1 & 0
\end{array}\right]$.
In the GPM objective, the reward model embeds the input text-image pair into a latent space, facilitating intricate preference modeling for text-to-image evaluation, which shows better performance in our experiments. With GPM preference embedding, LLaVA-Reward can produce more accurate preference without scalar rewarding, which benefits pairwise evaluations and direct preference optimization~\citep{rafailov2023direct}.  
LLaVA-Reward also provides BT-styled variants when scalar rewarding is necessary.
For unpaired human scoring or binary annotated data, such as the binary labeled NSFW image dataset~\citep{qu2023unsafe,Crone_2018} for text-to-image safety, \method{} is trained with Cross-Entropy (CE) loss, with the chosen (safe) and rejected (unsafe) images labeled as true and false: 
\begin{equation*}
\begin{aligned}
&\mathcal{L}_{\mathrm{CE}}= \mathcal{L}_{\mathrm{CE}}^{\vi_c} + \mathcal{L}_{\mathrm{CE}}^{\vi_r} \\[4pt]
  &= -\mathbb{E}_{(\vi_c,\vi_r,\vt)\sim\mathcal D_p}
     \Bigl[
       \log \sigma\!\bigl(s_{\theta_p}(\vi_c,\vt)\bigr)  +\log\bigl(1-\sigma\!\bigl(s_{\theta_p}(\vi_r,\vt)\bigr)\bigr)
     \Bigr]
\end{aligned}
\end{equation*}
where $s\left(\vi, \vt\right)=r_{\theta_p}(\vi, \vt)$, is the scalar output of \method{}. We label images cross-promptly according to scores in the case of unpaired human scoring data, given the insight from \cite{sun2024rethinking}. In both cases: $\mathcal{L}_{\mathrm{ranking}}$ or $\mathcal{L}_{\mathrm{CE}}$, when given paired data, we can utilize trained \method{} to compute the probability that image $\vi_c$ is preferred over $\vi_r$ given shared caption $\vt$ as: 
\begin{equation}
    \mathbb{P}\left(\vi_c \succ\vi_r \mid \vt\right)=\sigma(s(\vi_c, \vt)-s(\vi_r, \vt)).
\end{equation}
\vspace{-2em}
\paragraph{Training Details}
\method{} supports multiple types of data formats to perform preference learning, which enables it to include more dataset into the training. During the training, we freeze the visual encoder and the internal language model. Only the visual projector, the reward head $f_{\mathrm{SCA}}$ and the LoRA-adpter $\Delta \mathbf{W}$ of the language model are unfrozen for fine-tuning, with $~$8$\%$ additional parameters. 

%% file: sec/4_experiment.tex
\section{Experiments}
\method{} is finetuned for the evaluation of text-to-image generation in terms of text-image alignment, artifact, and safety. It can also serve as a reward model within diffusion inference-time scaling\cite{singhal2025general}, demonstrating great performance. All experiments are done using PyTorch on NVIDIA A6000 or A100 GPUs.
\input{tabel/mjbench}
\subsection{Baselines}
The baselines for the evaluation of text-to-image generations are categorized into three groups as follows:

\begin{itemize}
    \item \noindent\textbf{xLIP-based score methods}: HPS-v2.1~\citep{wu2023human}, BLIP-2~\citep{li2023blip}, PickScore-v1~\citep{kirstain2023pick}, ImageReward~\citep{xu2024imagereward}, Aesthetics~\citep{schuhmann2022laion}, CLIPScore\citep{hessel2021clipscore},
    \item \noindent\textbf{MLLMs}: InstructBLIP~\citep{dai2023instructblip}, MiniGPT4-v2~\citep{chen2023minigpt}, Qwen-VL-Chat~\citep{bai2023qwen}, Internvl-chat-v1-5~\citep{chen2024far}, Idefics2~\citep{laurenccon2024matters} and LLaVA~\citep{liu2024llavanext,liu2023improvedllava,liu2023llava}. We also include 4 closed-source MLLMs: GPT-4V, GPT-4o, Gemini-Ultra, and Claude-3-Opus.
    \item \noindent\textbf{MLLM-based reward models}: EVALALIGN~\citep{tan2024evalalign}, VQAScore~\citep{lin2024evaluating} and LLaVA-score~\citep{li2024removing} for text-image alignment, ImageGuard~\citep{li2025t2isafety} and LlavaGuard~\citep{helff2024llavaguard} for the safety evaluation of text-to-image generations.
\end{itemize}

\noindent For reward model baselines for diffusion inference time scaling, we include ImageReward, and CLIPscore.

\subsection{Datasets and Metrics}
We evaluate the reward/preference performance of \method{} and baselines from three perspectives: image-text alignment, fidelity, and safety. We conduct experiments to evaluate each of them on the MJ-Bench~\citep{chen2024mj} corresponding subset, reporting the accuracy with and without links of successfully ranking image pairs of the same prompt. Furthermore, for image-text alignment, we further evaluate TIFA 160 with Pearson, Kendall, and pairwise accuracy~\citep{deutsch2023ties} as metrics. For image safety evaluation, we additionally evaluate Unsafe Diffusion and SMID, and report the F1 score between predictions and binary labels. For experiments of diffusion inference time scaling, we sample image examples using the prompts of GenEval benchmark~\citep{ghosh2023geneval} and DrawBench~\citep{saharia2022photorealistic}. We evaluate image quality with \method{}, ImageReward, CLIPScore, HPSv2, VQAScore, GenEval\citep{ghosh2023geneval} and the LLM Grader prompting GPT-4o with the prompt from \citet{ma2025inference}.
\subsection{Training data}\label{sec:trainingdata}
\begin{table}[t!]
\caption{Training data statistics. ``IR" denotes for ImageReward.}
\vspace{-0.3cm}
\centering
\setlength{\tabcolsep}{5pt}
\begin{small}
\begin{tabular}{@{}l|c|c|c@{}}
\toprule
{Task} & {Source} & {Data size} & {Format} \\
\midrule
Text-image alignment           & IR-alignment &158k  & Pairwise \\
Artifact/fidelity                & IR-fidelity & 84k & Pairwise \\
Safety              & UnsafeBench & 8.1k  & Binary \\
Inference-time scaling       & IR-ranking & 136k & Pairwise \\
\bottomrule
\end{tabular}
\end{small}
\label{table:stat}
\end{table}
\input{tabel/safety_alignment}
We use ImageReward~\citep{xu2024imagereward} as the training set of \method{} for text-image alignment and the fidelity perspective by converting the corresponding human preference ratings into pairwise preference data. This model is then used for the diffusion-based inference-time scaling. 
For the image safety training set, we utilize the binary labeled UnsafeBench training set~\citep{qu2024unsafebench} to train \method{}. 

For the text-image alignment training set, we perform data filtering based on the Levenshtein distance~\citep{yujian2007normalized} between each prompt and those in the MJ-Bench alignment set to erase the overlap between them. We additionally create 70K preference pairs by introducing hard negative samples. Specifically, for each prompt $\vt$ and its best aligned images $\vi_k$, $k\in [1,m]$ as $m$ images that might have the same highest score, we find the prompt $\vt_\mathrm{neg}$ in ImageReward that is most similar to $\vt$ as the negative prompt and form $\{\vt_\mathrm{neg},\vi_k\}$ as the negative text-image pair of positive pair $\{\vt,\vi_k\}$. Similarly, for each best aligned image $\vi_k$ of prompt $\vt$, we find the most similar image $\vi_{k,\mathrm{neg}}$ as the negative image and form the negative pair as $\{\vt,\vi_{k,\mathrm{neg}}\}$ for $\{\vt,\vi_k\}$. In this way, we enrich the original training set with the hardest negative prompts and images, improving \method{}'s capacity to understand complex text-image relationships.
Detailed training data statistics are shown in \cref{table:stat}. The implementation details of \method{} are listed in \cref{app:detail}.

\begin{figure*}[t!]
    \centering
    \includegraphics[width=0.88\linewidth]{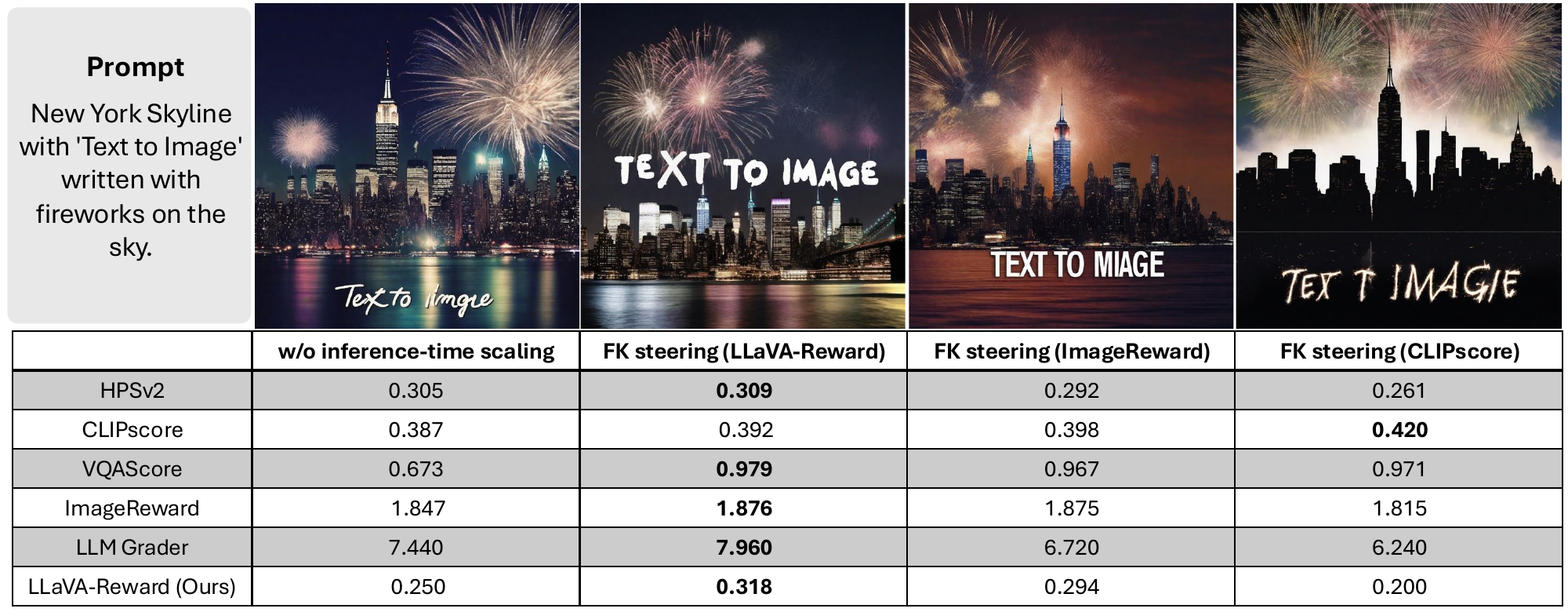}
    \vspace{-0.6em}
    \caption{Examples of diffusion inference-time scaling via FK steering (SDXL) using 4 different reward models with the prompt from DrawBench (top) and GenEval (bottom). The LLM grader is conducted using GPT-4o with prompts from \citet{ma2025inference}.}
    \label{fig:example_generation}
    \vspace{-1em}
\end{figure*}

\subsection{Performance on Text-to-image Evaluation }
For MJ-Bench results in \cref{tab:main_result}, we first compare \method{} with ImageReward which shares the same training data (we filtered 2k data in the alignment set). Compared with BLIP-based ImageReward, \method{} improved by about 31$\%$ to 43$\%$, which indicates the benefits of the MLLM backbone in generalization capacity for OOD data; For the MLLM VQA-based method: EVALALIGN, ImageGuard, and LlavaGuard in \cref{tab:main_result}, \cref{tab:align_results} and \cref{tab:safety_results}, although their scores/judgments are relatively good (still fall behind \method{}) for binary labeled datasets with clear boundary, \textit{e.g.}, TIFA 160, Unsafe Diffusion and SMID. But for text-to-image preference judgment for MJ-Bench, where the gaps between chosen and rejected samples are small,  MLLM VQA-based reward models are easily trapped into ties, for example, ImageGuard gives all tied preference on MJ-Bench safety set. Overall, \method{} outperforms these methods greatly. 

Compared to VQAScore and LLaVA-score in \cref{tab:main_result} and \cref{tab:align_results}, \method{} still shows the best results with much smaller model size. Besides, \method{} is more flexible and can be easily customized for evaluating different perspectives with the switch of training data. Comparing \method{} with or without the SkipCA module (we replace the SkipCA module with a single-layer MLP as the reward head for ``w/o SkipCA"), we can observe a performance gap, especially in MJ-Bench safety set, which indicates the importance of revisiting the diluted visual tokens in the later stage for text-to-image rewarding for decoder-only MLLMs; \textbf{5)} In general, compared with CLIP/BLIP based methods and general MLLMs, \method{} shows outperformed results via efficient and flexible LoRA adaptation. It achieves a balance between performance and scale versus closed-source models.
\begin{table}[t!]
\caption{Average inference time per evaluation for MLLM-based methods for text-to-image evaluation.}
\vspace{-0.3cm}
\centering
\begin{small}
\setlength{\tabcolsep}{5pt}
\begin{tabular}{@{}lc|c@{}}
\toprule
\textbf{Method} & \textbf{Type} & \textbf{Inference Time (s)} \\
\midrule
ImageGuard&VQA& 0.85\\
LlavaGuard&VQA& 4.30 \\
Evalalign&VQA& 7.01  \\
\midrule
LLaVA-score &Token& 0.26 \\
VQAScore &Token& 2.81 \\
\midrule
\method{} & Hidden Embed. & 0.35 \\
\bottomrule
\end{tabular}
\end{small}
\label{table:time}
\vspace{-0.3cm}
\end{table}
\vspace{-0.3cm}
 \paragraph{Inference time comparison.} In \cref{table:time}, we compare the average inference time per evaluation between the \method{} and MLLM-based baselines. \method{} achieves the second-best efficiency, much faster than VQA-based methods. In general, VQA-based methods are more time-consuming, due to detailed system prompts and the response manner. Evalalign and LlavaGuard will also produce a long rationale after the evaluation text. This lengthy rationale is meaningless if the evaluation is inaccurate and introduces only heavy computation costs. Token likelihood-based methods have shorter average inference because they only need to extract the likelihood of a few tokes after instructions. Similarly to \method{}, LlavaScore utilizes short instructions that contribute to its high efficiency.
\subsection{Improving Text-to-Image Generation}
\method{} achieved outstanding performance in evaluating text-to-image generation, which implies its potential to improve text-to-image generation. Given the recent success in diffusion inference-time scaling~\cite{ma2025inference,singhal2025general}, we apply \method{} as the reward model in diffusion inference-time scaling, Feynman Kac steering (FK steering)~\citep{singhal2025general} in specific. We show that \method{} can achieve a greater performance gain in the generations of Stable Diffusion v2.1 and Stable Diffusion XL~\cite{rombach2022high, podell2023sdxl} than other text-to-image reward models.
\vspace{-0.4cm}
\paragraph{Diffusion inference-time scaling in Text-to-Image.} Diffusion inference-time scaling is a training-free method during diffusion denoising sampling that selects the better intermediate noise~\citep{ma2025inference} or generations~\cite{singhal2025general} and resumes the sample from the selected intermediate productions without increasing the NFE. In our experiments, we use FK steering as our inference-time scaling method, which applies sequential Monte Carlo (SMC) during inference-time steering. We perform experiments on the prompts from the GenEval benchmark~\citep{ghosh2023geneval} and DrawBench~\citep{saharia2022photorealistic} with different reward models: \method{}, ImageReward, and CLIPScore. We set the particle number as 4 for all the experiments. The $\lambda$ for \method{}, ImageReward, and CLIPscore are setted as 60, 10 and 20, respectively. 

In \cref{fig:example_generation}, we can observe that FK steering with \method{} can achieve the best prompt alignment towards the ``purple train" and ``brown own" compared to other reward models. Additionally, for the DrawBench prompt that includes text, only the generation using \method{} as the reward model displays the correct text "Text to Image". We include more visual results in \cref{app:visual}.

In \cref{tab:generation} of \cref{app:visual}, we show the overall performance of 3 reward models in GenEval and DrawBench using SD v2.1 and SDXL, and \cref{tab:geneval} shows the GenEval scores on different GenEval tasks. We can observe that each diffusion inference-time scaling variant tends to have a better result when evaluated on itself, such as the CLIPScore variant's performance on GenEval using SD v2.1 on its own. However, in most cases, except for ImageReward, \method{} achieves the best performance, indicating that \method{} can capture more general text-to-image rewards and characters. Compared to ImageReward, which shares the same training data, \method{} outperforms it in all unoverlapped metrics, especially for VQAScore, which targets alignment evaluation. It demonstrates that \method{} can improve the overall quality of the generation and also the image alignment with the prompts with complex text-image relationships in GenEval and DrawBench.
\subsection{Ablation Studies}
\paragraph{Affect of different objectives.} In \cref{tab:abla_objective}, we compare \method{} results with BT loss or 2-dimensional GPM loss on the text-image alignment set of MJ-Bench. We can observe that models with 2-dimensional GPM loss can provide slightly better evaluation results less than 1$\%$. GPM loss can facilitate better pairwise preference production. However, because the model fine-tuned with GPM loss cannot be used for reward generation, BT loss is a more general and balanced choice.
\vspace{-0.2cm}
\paragraph{Impact of the negative samples in training data.}
We investigate the effect of augmented text-image alignment training data introduced in \cref{sec:trainingdata}. As shown in \cref{tab:abla_objective}, the performance of ``\method{} w/o Neg", which is fine-tuned without extra 70k pairs with negative images or prompts, drops to 1.6$\%$. It shows that the extra preference data with hard negative samples can help the model to identify the difference between very similar positive and negative prompts or images, thus judging the alignment between images and texts more precisely.
\begin{figure}[t]
    \centering
    \begin{minipage}[t]{0.48\linewidth}
        \centering
        \includegraphics[width=\linewidth]{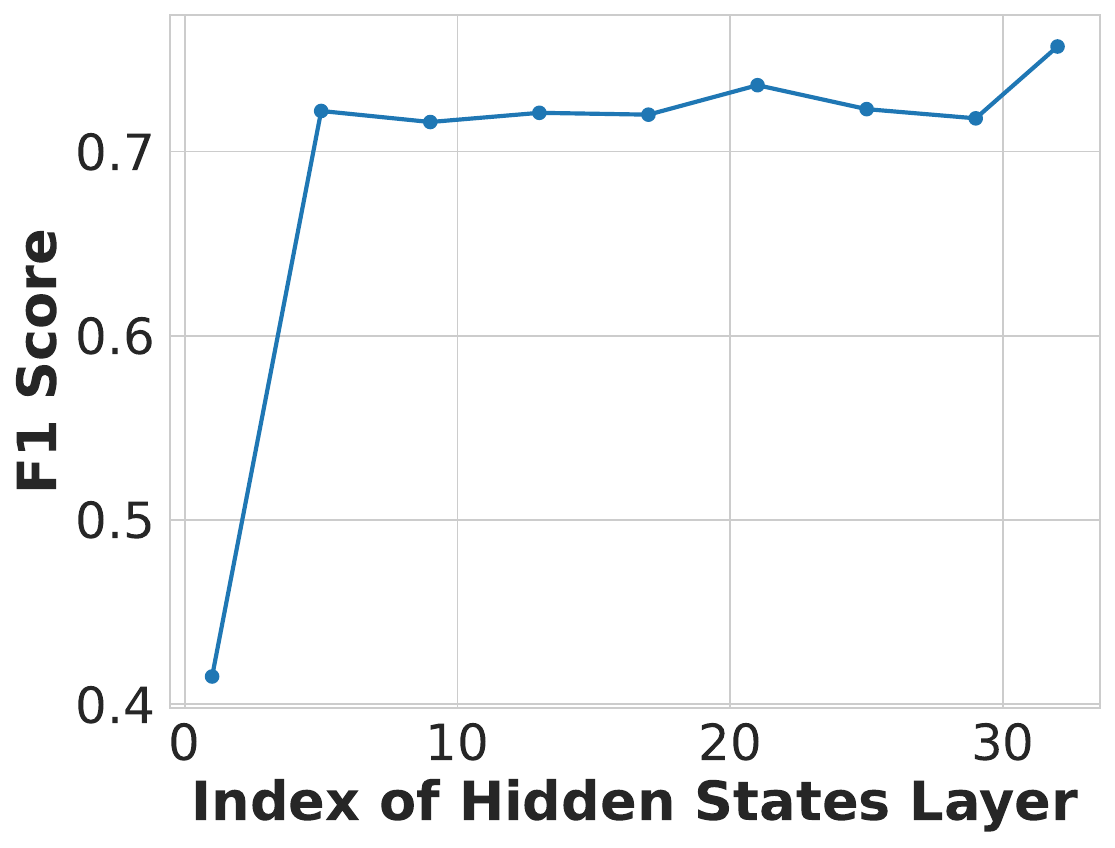}
        \vspace{-0.8em}
        \end{minipage}
    \hfill
    \begin{minipage}[t]{0.48\linewidth}
        \centering
        \includegraphics[width=\linewidth]{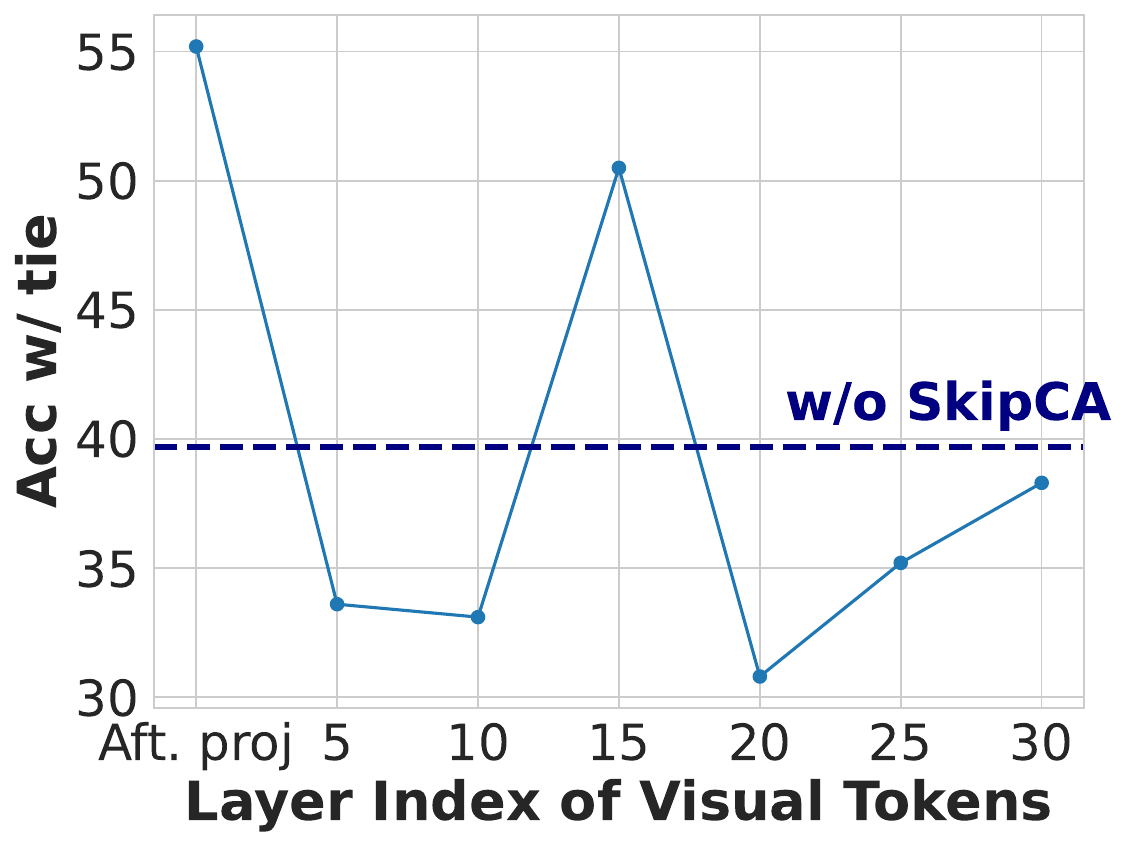}
    \end{minipage}
    \vspace{-0.3cm}
    \caption{Ablation Results. Left: F1 score on safety benchmark, SMID,  using different layers of hidden state. Right: Accuracy on MJ-Bench with different layers of visual tokens in SkipCA.}
    \label{fig:abla}
    \vspace{-0.5cm}
\end{figure}
\vspace{-0.2cm}
\paragraph{Hidden state extraction manners.} In \cref{tab:abla_objective}, we also report ``\method{} with Mean", the results of extracting hidden state by averaging over all tokens instead of using only the EOS token for model outputs. We can see that ``\method{} with Mean" is barely better than random choice. It demonstrates that the hidden state at the EOS token already has enough information for human preference modeling and fits our model better.
\vspace{-0.2cm}
\paragraph{Model size and training paradigms.} We construct \method{} with different model sizes: 8.3B variant with Qwen2.5-VL~\citep{bai2025qwen2} and 13.4B variant with LLaVA-NeXT~\citep{liu2024llava}. Their results on MJ-Bench's image fidelity evaluation are shown in \cref{tab:abla_size}. Due to the stronger visual encoder with the better visual recognition capacity in Qwen2.5-VL, the Qwen2.5-VL variant has a 3.1$\%$ improvement compared to Phi-3.5-vision variant. In addition, fully fine-tuning Phi-3.5 vision with fixed visual encoder also leads to a performance loss at 4.8$\%$. Moreover, it increases the training time by 15 hours (+110$\%$). To ensure better efficiency of the reward model, we use the Phi-3.5 vision variant for our main experiments.
\vspace{-0.2cm}
\paragraph{Choice of layer for the hidden state.}   
In \method{}, we feed the last hidden state representation (layer 32) into the reward head to output the reward. However, hidden states from other layers should also be considered, as layers differ in the extracted features~\citep{ben2024attend}. Recent MLLM-based document retrieval methods, such as LoCAL~\citep{chen2024lora}, found that the use of hidden states from earlier layers for retrieval can have comparable or even better performance with better computational efficiency. Thus, we also conduct an ablation study to investigate the impact of using different layers of hidden state for \method{}. Specifically, we conduct experiments on the safety evalution with different hidden state layer choices and show the F1-score results on SMID in \cref{fig:abla}. We can observe that using the hidden state of the middle layer (21) can lead to a relatively better result but still falls behind the last layer. Therefore, we maintain the setting that the hidden state of the last layer (32) is used for the reward head.
\vspace{-0.2cm}
\paragraph{Impact of visual tokens from different layers.} In SkipCA module, we extract the visual embedding after the visual projector because visual-modal feature in visual token will dilute in the deeper transformer layers. We demonstrate our design here by diversifying the layer of visual tokens used in SkipCA. In \cref{fig:abla}, we show the safety evaluation performance on MJ-Bench. It is can be observed that the accuracy after of using visual tokens after the visual projector significantly better than other layers, which supports our choice in SkipCA.

%% file: tabel/mjbench.tex
\begin{table*}[t!]
    \vspace{-0.2in}
    \centering
    \caption{Evaluation of multimodal reward/score models across three perspectives, alignment, safety, and artifact in \textbf{MJ-Bench}. Five sets of methods are listed from top to bottom: CLIP-based models, open-source MLLMs, closed-source MLLMs, MLLM-based models, and \method{}.
    \textbf{Bold}: best; \underline{Underline}: second best; \textbf{\textsl{slanted}}: best results of closed-source LMMs; $\dagger$: results obtained from previous work~\citep{chen2024mj}; Cells in the same color indicate the model with the same training set.}
    \setlength{\tabcolsep}{2pt}
    \renewcommand{\arraystretch}{0.9}
    \begin{small}
    \begin{tabular}{lc|cc|cc|cc}
    \toprule
         \multirow{2}{*}{method}  &\multirow{2}{*}{$\#$Param}& \multicolumn{2}{c|}{\bf Alignment} & \multicolumn{2}{c|}{\bf Safety} & \multicolumn{2}{c}{\bf Artifact (Fidelity)} \\
         && Acc w/ tie & Acc w/o tie & Acc w/ tie & Acc w/o tie & Acc w/ tie & Acc w/o tie \\
    \midrule
         CLIPScore$\dagger$ &428M& $38.1$ & $59.5$ & $12.7$ & $33.3$ & $34.4$ & $68.4$ \\
         BLIP-2$\dagger$ &3.8B &$17.3$ & $38.8$ & $44.0$ & $65.6$ & $7.5$  & $36.5$ \\
         PickScore-v1$\dagger$&986M &$58.8$ & $64.6$ & 37.2 & $42.2$ & $83.8$ & $89.6$ \\
         HPS-v2.1$\dagger$ &2B &$47.3$ & $\bf 70.1$ & $18.8$ & $41.3$ & $67.3$ & $\underline{93.5}$ \\
         ImageReward$\dagger$ &478M &\cellcolor{gray!30}$50.9$ &\cellcolor{gray!30} $64.7$ & $24.9$ & $38.7$ &\cellcolor{gray!30} $63.5$ &\cellcolor{gray!30} $81.8$  \\
         Aesthetics$\dagger$  &304M &$32.4$ & $52.7$ & $27.0$ & $53.6$ & $69.6$ & $92.5$  \\
    \midrule
         LLaVA-1.5$\dagger$ &7B &$22.0$ & $50.8$ & $24.8$ & $50.2$ & $12.4$ & $51.6$  \\
         LLaVA-1.5$\dagger$ & 13B&$10.3$ & $51.9$ & $30.7$ & $60.7$ & $23.3$ & $61.2$ \\
         LLaVA-1.6-mistral$\dagger$ &7B& $31.3$ & $62.7$ & $15.2$ & $40.9$ & $45.8$ & $73.2$ \\
         LLaVA-1.6-vicuna$\dagger$ & 13B&$29.1$ & $60.3$ & $27.9$ & $45.6$ & $36.8$ & $62.5$  \\
         InstructBLIP$\dagger$ &7B& $17.1$ & $49.8$ & $26.4$ & $46.9$ & $25.2$ & $64.1$  \\
         MiniGPT4-v2$\dagger$ &7B&$32.8$ & $51.2$ & $25.7$ & $60.1$ & $36.7$ & $47.8$ \\
         Qwen-VL-Chat$\dagger$       &7B&$52.1$ & $31.6$ & $26.8$  & $7.1$  & $23.6$ & $24.6$ \\
         Internvl-chat-v1-5$\dagger$ & 26B&$55.3$ & $67.6$ & $6.3$   & $60.0$ & $66.3$ & $65.1$  \\
         Idefics2$\dagger$       & 8B &$32.6$ & $43.5$ & $13.6$  & $52.0$ & $46.1$ & $68.9$ \\
    \midrule
         GPT-4-vision$\dagger$ & -&$66.1$ & $67.0$ & $26.5$ & $97.6$ & $90.4$ & $96.5$  \\
         GPT-4o$\dagger$ &- &$61.5$ & $62.5$ & \bf\textsl{{35.3}} & \bf\textsl{{100.0}} & \bf\textsl{{97.6}} & \bf\textsl{{98.7}} \\
         Gemini Ultra$\dagger$ &- & \bf$\textsl{67.2}$ & \bf\textsl{69.0} & $13.1$ & $95.1$ & $55.7$ & $96.7$ \\
         Claude 3 Opus$\dagger$ &-& $57.1$ & $55.9$ & $13.4$ & $78.9$ & $11.9$ & $70.4$  \\
    \midrule
        EVALALIGN &13B &$11.7$ & $64.6$ & - & - & $54.4$ & $81.9$  \\
        VQAScore & 11B&$63.2$ & $63.8$ & - & - & - & -  \\
        LLaVA-score & 13B&$\underline{64.2}$ & $64.5$ & - & - & - & -  \\
        ImageGuard &7B& - & - & $0.0$ & $0.0$ & - & -  \\
        LlavaGuard &7B& - & - & $5.6$ & $\underline{90.9}$ & - & -  \\
    \midrule
         \method{}-Phi w/o  SkipCM & 4.2B& $\cellcolor{gray!30} \bf68.2$ & \cellcolor{gray!30}$\underline{68.6}$ & $39.7$ & $81.1$ & \cellcolor{gray!30}$87.3$ & \cellcolor{gray!30}$87.7$ \\
        \method{}-Phi &4.2B& $\cellcolor{gray!30} 66.1$ & $\cellcolor{gray!30}66.2$ & $ \underline{55.2}$ & $ \underline{92.1}$ & \cellcolor{gray!30}$\underline{91.1}$ & \cellcolor{gray!30}$91.2$  \\
        \method{}-Qwen &8.2B& $\cellcolor{gray!30} \underline{67.5}$ & $\cellcolor{gray!30}67.5$ & $\bf 59.2$ & $\bf 93.6$ & \cellcolor{gray!30}$\bf94.3$ & \cellcolor{gray!30}$\bf94.3$  \\
    \bottomrule
    \end{tabular}%
    \end{small}
    \label{tab:main_result}
    \vspace{-0.3cm}
\end{table*}

\begin{table}[t!]
    \centering
    \caption{Evaluation on image-text \textbf{alignment} performance of recent LMM-based model on TIFA 160. \textbf{Bold}: best; \underline{Underline}: second best.}
    \vspace{-0.3cm}
    \setlength{\tabcolsep}{2pt}
    \renewcommand{\arraystretch}{0.9}
    \begin{small}
    \begin{tabular}{l|ccc}
    \toprule
         \multirow{2}{*}{method}& \multicolumn{3}{c}{\bf TIFA 160}  \\
         & Pearson~$\uparrow$ & Kendall~$\uparrow$ & Pairwise Acc~$\uparrow$ \\
    \midrule
        EVALALIGN-13B & $46.7$ & $39.0$ & $36.5$   \\
        VQAScore-11B  & $66.9$ & $\bf52.9$ & $\bf71.6$ \\
        LLaVA-score-13B & $66.6$ & $51.5$ & $70.9$\\
    \midrule
        \method{}-4.2B & \bf71.1 & \underline{52.2} & \underline{71.3}  \\
        \method{}-8.2B & \underline{68.4}  & 49.4 & 70.0  \\
    \bottomrule
    \end{tabular}%
    \end{small}
    \label{tab:align_results}
    \vspace{-0.2cm}
\end{table}
\begin{table}[t!]
    \centering
    \caption{Evaluation of recent LMM-based \textbf{safety} judges on TIFA 160. \textbf{Bold}: best; \underline{Underline}: second best.}
    \vspace{-0.3cm}
    \setlength{\tabcolsep}{2pt}
    \begin{small}
    \begin{tabular}{l|cc}
    \toprule
        method& \bf UnsafeDiff & \bf SMID \\
    \midrule
        ImageGuard-7B & $84.1$ & $73.4$   \\
        LlavaGuard-7B & $84.5$ & $69.5$ \\
    \midrule
        \method{}-4.2B & \bf87.2 & \underline{75.7}  \\
        \method{}-8.2B & \underline{86.4} & \bf78.1  \\
    \bottomrule
    \end{tabular}%
    \vspace{-0.3cm}
    \end{small}
    \label{tab:safety_results}
\end{table}
\begin{table*}[t!]
    \vspace{-0.3in}
    \centering
    \caption{Diffusion inference-time scaling results of FK steering~\citep{singhal2025general} using the prompts of GenEval. We report the mean GenEval scores of different GenEval tasks. \textbf{Bold}: best; \underline{Underline}: second best.}
    \vspace{-0.2cm}
    \setlength{\tabcolsep}{3pt}
    \renewcommand{\arraystretch}{0.9}
    \begin{small}
    \begin{tabular}{lcc|c|c|c|c|c|c|c}
\toprule
\multirow{2}{*}{Dataset} &
\multirow{2}{*}{Reward Model} &
\multirow{2}{*}{Model} &
\makecell[c]{Attribute} &         
\multirow{2}{*}{Counting} &
\multirow{2}{*}{Position} &
\multirow{2}{*}{Colors} &
\makecell[c]{Two} &
\makecell[c]{Single} &
\multirow{2}{*}{Overall} \\      

& & & binding & & & & Objects & Object & \\        
\midrule
       \multirow{8}{*}{GenEval} & None &\multirow{4}{*}{SD v2.1}&0.165&0.506&0.103&0.848&0.525&0.978&0.521\\
        & CLIPScore &&0.220&0.497&0.103&\underline{0.880}&0.689&0.984&0.562\\
         &ImageReward &&\underline{0.327}&\bf0.631&\underline{0.105}&0.851&\bf0.765&\bf0.997&\underline{0.613}\\
        &\method{} &&\bf0.375&\underline{0.619}&\bf0.108&\bf0.920&\underline{0.719}&\underline{0.991}&\bf0.622\\
        \cmidrule{2-10}
        &None &\multirow{4}{*}{SDXL}&0.215&0.431&0.135&0.862&0.742&0.991&0.563\\
         &CLIPScore &&0.238&0.519&0.078&\underline{0.883}&0.833&\bf1.000&0.592\\
         &ImageReward &&\underline{0.280}&\underline{0.581}&\bf0.177&0.869&0.851&\bf1.000&\underline{0.627}\\
        &\method{} & &\bf0.313&\bf0.672&\underline{0.162}&\bf0.894&\underline{0.836}&0.991&\bf0.645\\
    \bottomrule
    \end{tabular}%
    \vspace{-0.3cm}
    \end{small}
    \label{tab:geneval}
\end{table*}

%% file: tabel/safety_alignment.tex
\begin{table}[t!]
    \centering
    \caption{Ablation Study results on fidelity in MJ-Bench of \method{} with different base models and training paradigms.}
    \vspace{-0.3cm}
    \setlength{\tabcolsep}{2pt}
    \renewcommand{\arraystretch}{0.9}
    \begin{small}
    \begin{tabular}{lcc|cc}
    \toprule
         \multirow{2}{*}{Backbone}&\multirow{2}{*}{$\#$Param} & \multirow{2}{*}{Tuning}  & \multicolumn{2}{c}{\bf Artifact (Fidelity)} \\
         &&& Acc w/ tie & Acc w/o tie \\
    \midrule
         Phi-3.5-vision& 4.2B& LoRA &  $91.1$ & $91.2$ \\
         Phi-3.5-vision& 4.2B& Full FT & $86.3$ & $87.0$ \\
         Qwen2.5-VL& 8.3B& LoRA & $94.3$ & $94.3$ \\
         LLaVA-NeXT& 13.4B& LoRA & $90.8$ & $91.1$ \\
 
    \bottomrule
    \end{tabular}%
    \end{small}
    \label{tab:abla_size}
\end{table}
\vspace{0.5em}
\begin{table}[t!]
    \centering
    \caption{Ablation Study results on text-image alignment in MJ-Bench of \method{} with different objectives, w/ or w/o negative samples, with different hidden state extraction manner.}
    \vspace{-0.3cm}
    \setlength{\tabcolsep}{2pt}
    \renewcommand{\arraystretch}{0.9}
    \begin{small}
    \begin{tabular}{lc|cc}
    \toprule
         \multirow{2}{*}{Method}&\multirow{2}{*}{Objective}  & \multicolumn{2}{c}{\bf Alignment} \\
         && Acc w/ tie & Acc w/o tie \\
    \midrule
         \method{} & BT& $65.2$ & $65.6$ \\
         \method{} w/o Neg& GPM& $64.5$ & $65.0$ \\
         \method{} with Mean& GPM& $54.6$ & $54.9$ \\
         \midrule
         \method{} & GPM& $66.1$ & $66.2$ \\
    \bottomrule
    \end{tabular}%
    \vspace{-0.3cm}
    \end{small}
    \label{tab:abla_objective}
\end{table}

%% file: sec/5_conclusion.tex
\vspace{-0.1cm}
\section{Conclusion}
We introduced \method{}, a multimodal reward model for text-to-image generation that takes advantage of MLLM's hidden embeddings for efficient human-aligned scoring. By integrating LoRA adapters and the SkipCA module, our approach enhances text-image reasoning while maintaining impressive computational efficiency. Empirical results show state-of-the-art performance in alignment, fidelity, and safety evaluations. Future work includes joint training of \method{} in multiple perspectives to develop a more universal reward model and improving performance on very complex evaluation tasks.

\paragraph{Acknowledgement.} This work is partially supported by NSF AI Institute-2229873, NSF RI-2223292, an Amazon research award, and an Adobe gift fund. Any opinions, findings and conclusions or recommendations expressed in this material are those of the author(s) and do not necessarily reflect the views of the National Science Foundation, the Institute of Education Sciences, or the U.S. Department of Education.

%% file: sec/X_suppl.tex
\clearpage
\setcounter{page}{1}
\maketitlesupplementary
\appendix

\section{Implementation Details}\label{app:detail}
We fine-tune Phi-3.5-vision on the training set introduced above via the standard pairwise ranking loss or 2-dimensional GPM~\citep{zhang2024general} loss when only preference is necessary, such as the evaluation in MJ-Bench. We finetune the safety model on UnsafeBench via cross-entropy loss. We train \method{} with batch size 8 and gradient accumulation size 4 on 4 NVIDIA A6000 GPUs for one epoch, with a learning rate of 2e-4. We fine-tuned the LoRA adapter with rank 128, the visual projector, and the SkipCA value head described in Section 3, with other parameters frozen.

\section{Limitations}
Multimodal LLMs have shown impressive performance in image understanding and multimodal reasoning. However, the multimodal reward model still suffers from the absence of high-quality training data, which can lead to issues such as reward hacking~\citep{skalse2025definingcharacterizingrewardhacking}.
Currently, \method{} is adapted using preference data derived from generations of a limited set of models. In future work, we aim to enhance the robustness and capacity of \method{} by incorporating more comprehensive and diverse training data.
\begin{table*}[b]
    \vspace{-0.2in}
    \centering
    \caption{Diffusion inference-time scaling results of FK steering~\citep{singhal2025general} using the prompts of GenEval and DrawBench. We report the mean performance on each metrics. 
    We use \textbf{bold} and \underline{underline} to indicate the best and second-best results.} 
    \vspace{-0.2cm}
    \setlength{\tabcolsep}{2pt}
    \renewcommand{\arraystretch}{0.9}
    \begin{small}
    \begin{tabular}{lcc|c|c|c|c|c}
    \toprule
         \multirow{1}{*}{Dataset} &\multirow{1}{*}{Reward Model} &\multirow{1}{*}{Model}& \multicolumn{1}{c|}{\bf HPSv2}&\multicolumn{1}{c|}{\bf ClipScore}&\multicolumn{1}{c|}{\bf VQAscore} & \multicolumn{1}{c|}{\bf \method{}} & \multicolumn{1}{c}{\bf ImageReward} \\
    \midrule
       \multirow{8}{*}{GenEval} & None &\multirow{4}{*}{SD v2.1}& $0.285$  & $0.300$  &$0.620$  &$-0.045$&$0.303$ \\
        & CLIPScore && $0.292$ & $\bf0.314$ &$0.663$  &$0.018$&$0.660$ \\
         &ImageReward && $\underline{0.298}$& $\underline{0.312}$ &$\underline{0.737}$&$\underline{0.078}$&$\bf1.179$ \\
        &\method{} && $\bf0.301$ & $0.311$ &$\bf0.743$ &$\bf0.226$&$\underline{0.976}$\\
        \cmidrule{2-8}
        &None &\multirow{4}{*}{SDXL}& $0.307$  & $0.303$  &$0.681$  &$0.180$&$0.677$ \\
         &CLIPScore && $0.316$ & $0.317$ &$0.694$  &$0.252$&$1.046$ \\
         &ImageReward && $\underline{0.316}$& $\underline{0.316}$ &$\underline{0.718}$&$\underline{0.270}$&$\bf1.247$ \\
        &\method{} & & $\bf0.323$ & $\bf0.319$ &$\bf0.721$&$\bf0.397$&$\underline{1.073}$ \\
        \midrule
        \multirow{8}{*}{DrawBench} & None &\multirow{4}{*}{SD v2.1}& $0.261$  & $0.296$  &$0.649$  &$0.023$&$0.849$ \\
        & CLIPScore && $0.267$ & $\underline{0.314}$ &$\underline{0.706}$  &$0.070$&$1.042$ \\
         &ImageReward && $\underline{0.274}$& $0.312$ &$0.702$&$\underline{0.124}$&$\bf1.332$ \\
        &\method{} && $\bf0.276$ & $\bf0.321$ &$\bf0.729$ &$\bf0.192$&$\underline{1.147}$ \\
        \cmidrule{2-8}
        &None &\multirow{4}{*}{SDXL}& $0.276$  & $0.327$  &$0.826$  &$0.244$&$1.309$ \\
         &CLIPScore && $0.282$ & $\underline{0.348}$ &$\bf0.890$  &$0.308$&$1.540$ \\
         &ImageReward && $\underline{0.286}$& $0.343$ &$0.881$&$\underline{0.299}$&$\bf1.627$\\
        &\method{} & & $\bf0.288$ & $\bf0.350$ &$\underline{0.887}$ &$\bf0.352$&$\underline{1.542}$ \\
    \bottomrule
    \end{tabular}%
    \vspace{-0.3cm}
    \end{small}
    \label{tab:generation}
\end{table*}
\newpage
\section{Additional Quantitative and Visual Results}\label{app:visual}
\begin{figure*}[t!]
    \centering
    \includegraphics[width=0.95\linewidth]{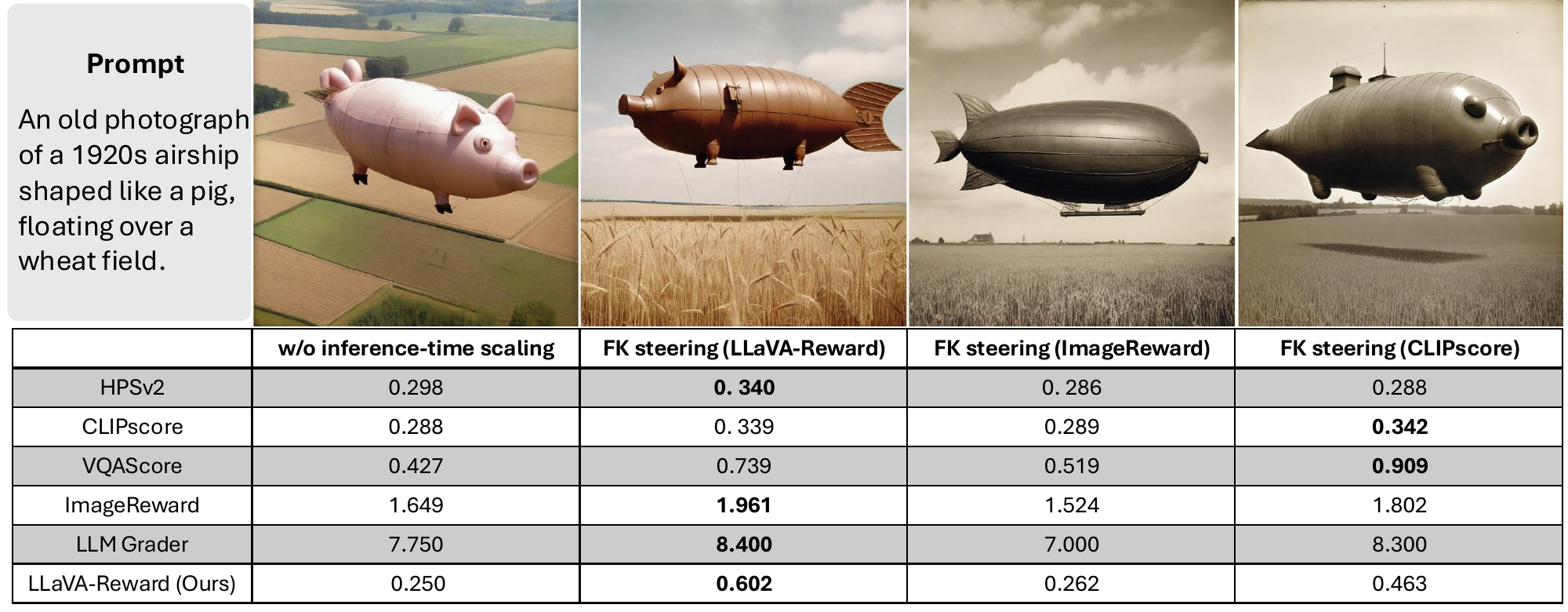}
    \includegraphics[width=0.95\linewidth]{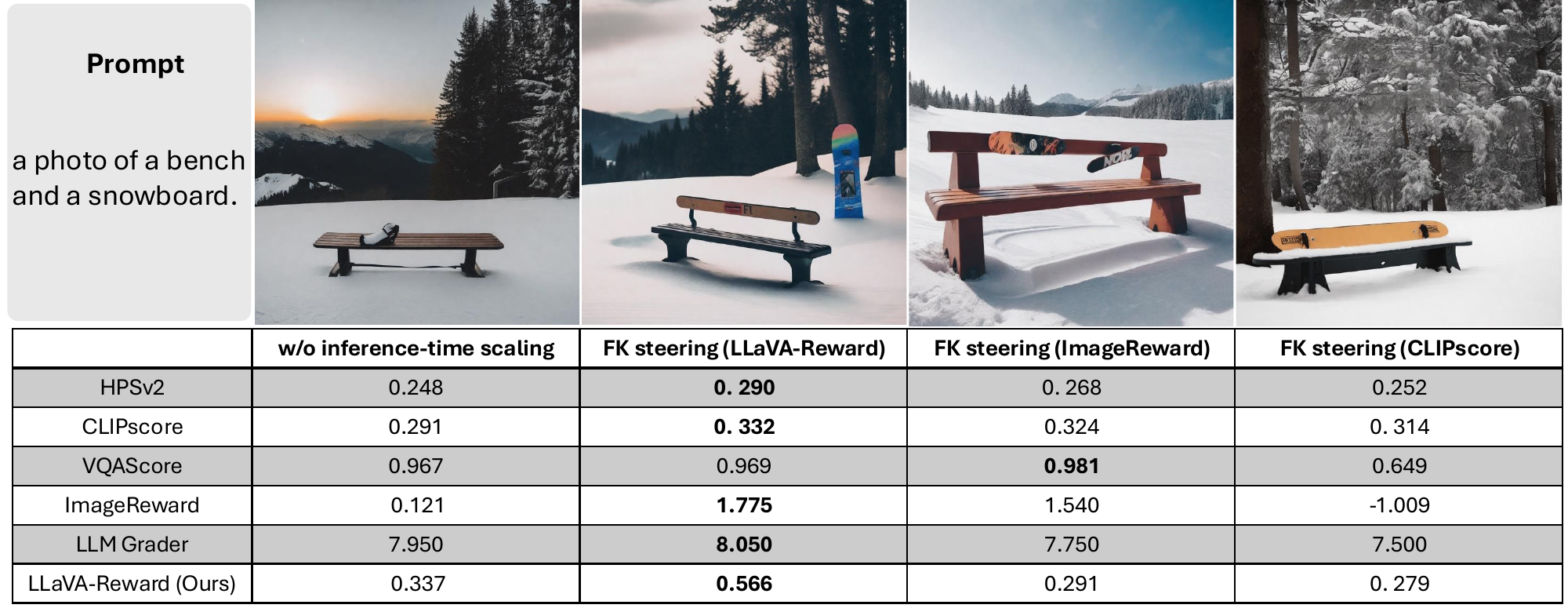}
    \includegraphics[width=0.95\linewidth]{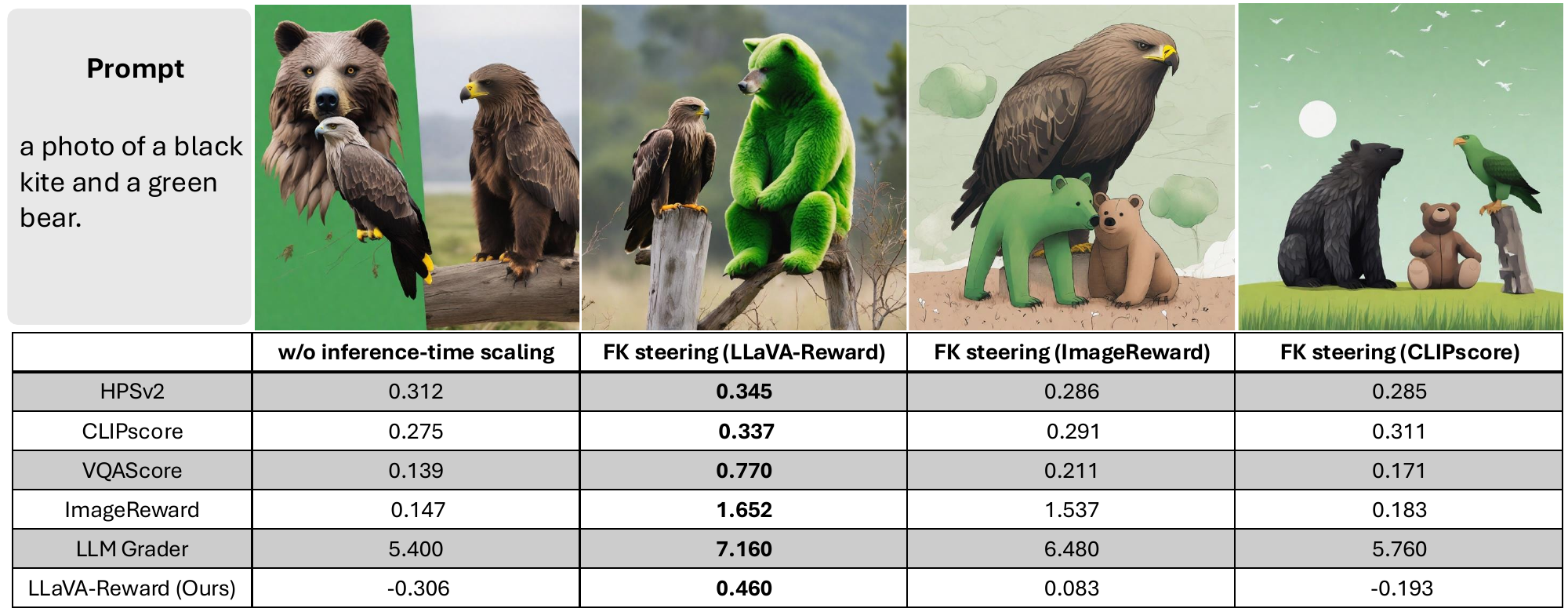}
    \vspace{-1em}
    \caption{Examples of diffusion inference-time scaling via FK steering (SDXL) using 5 different reward models with the prompt from GenEval. The LLM grader is conducted using GPT-4o with prompts from \citet{ma2025inference}.}
    \label{fig:sample0}
    \vspace{-1.5em}
\end{figure*}

\begin{figure*}[t!]
    \centering
    \includegraphics[width=0.95\linewidth]{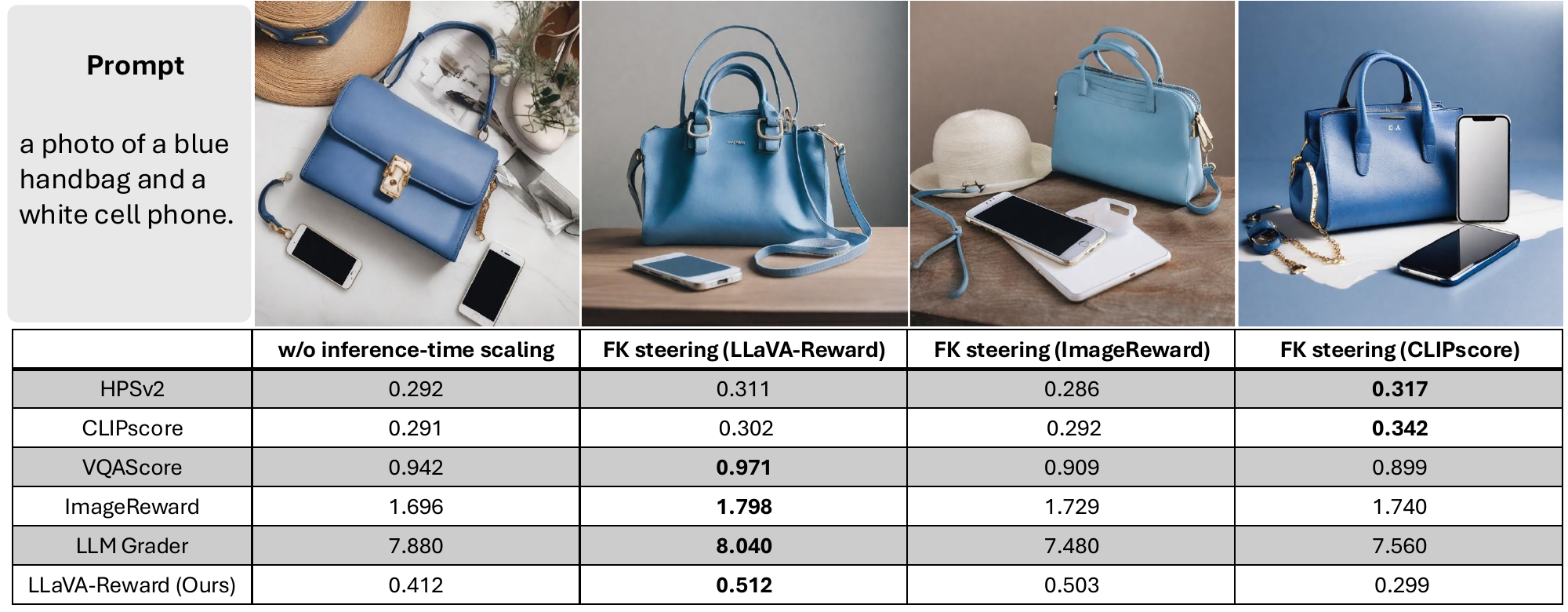}
    \includegraphics[width=0.95\linewidth]{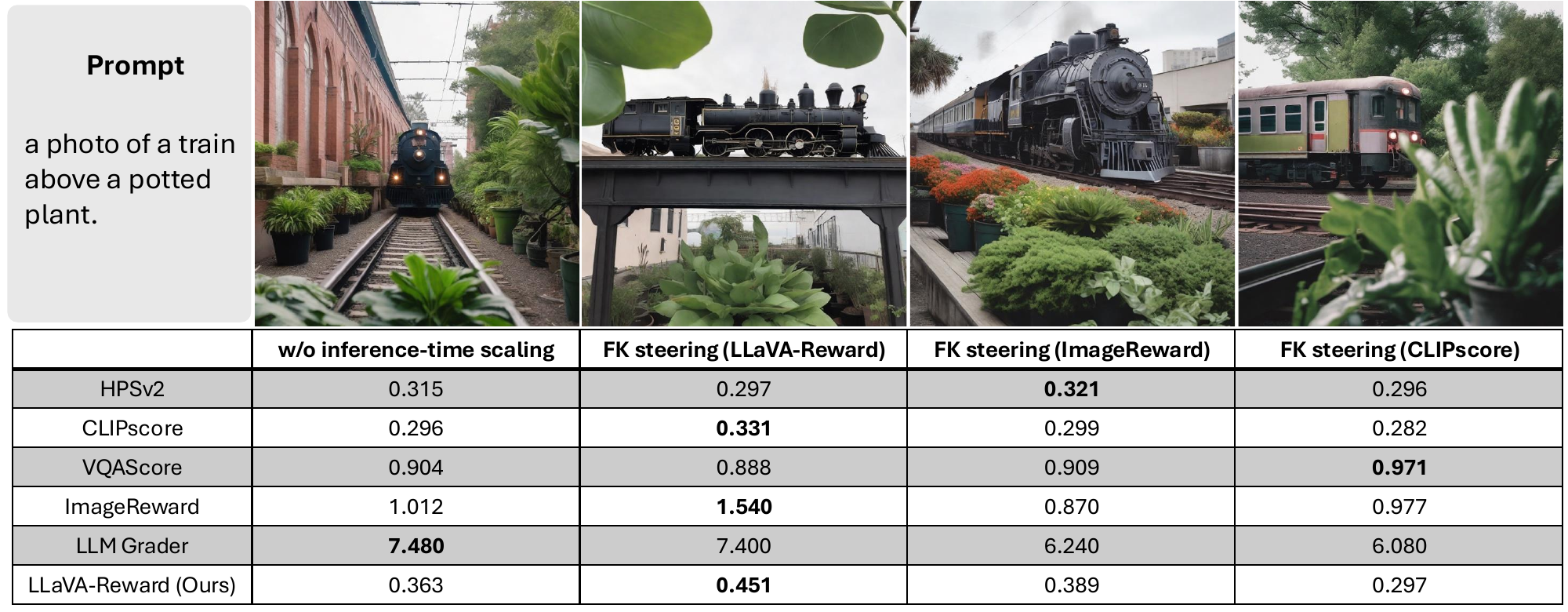}
    \includegraphics[width=0.95\linewidth]{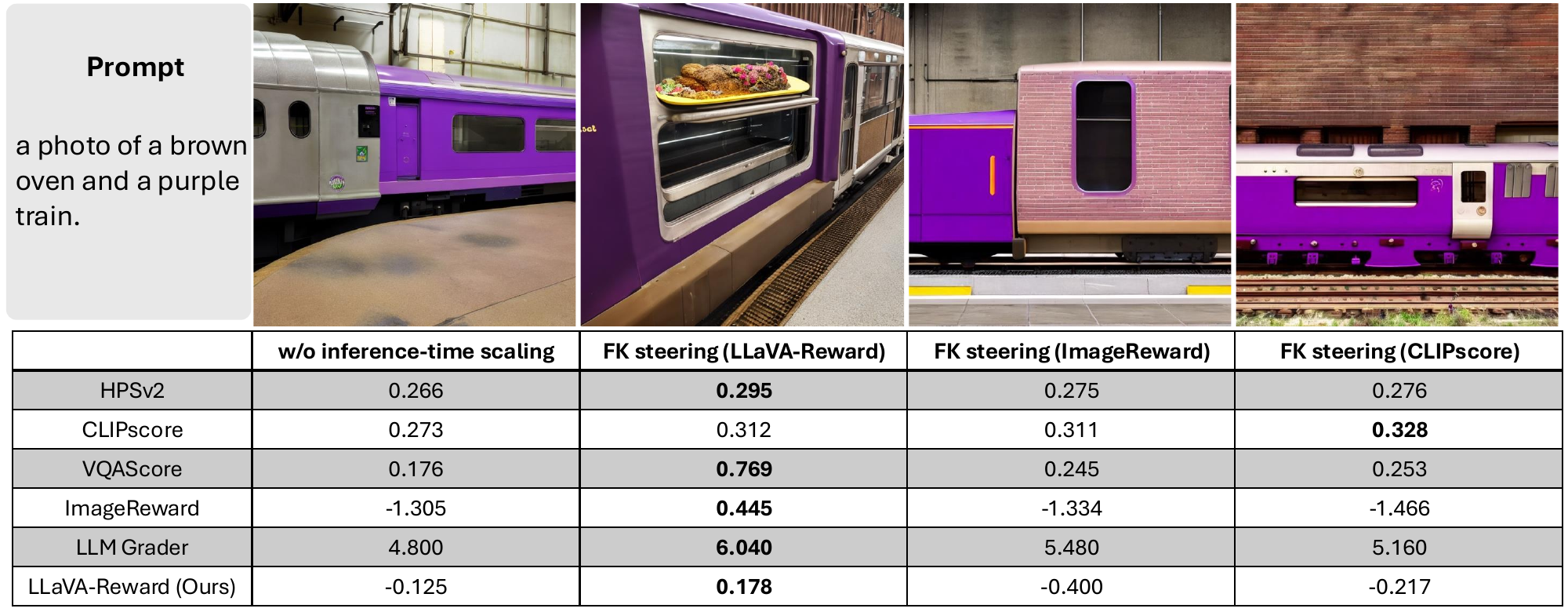}
    \vspace{-1em}
    \caption{Examples of diffusion inference-time scaling via FK steering (SDXL) using 5 different reward models with the prompt from GenEval. The LLM grader is conducted using GPT-4o with prompts from \citet{ma2025inference}.}
    \label{fig:sample1}
    \vspace{-1.5em}
\end{figure*}

\begin{figure*}[t!]
    \centering
    \includegraphics[width=0.95\linewidth]{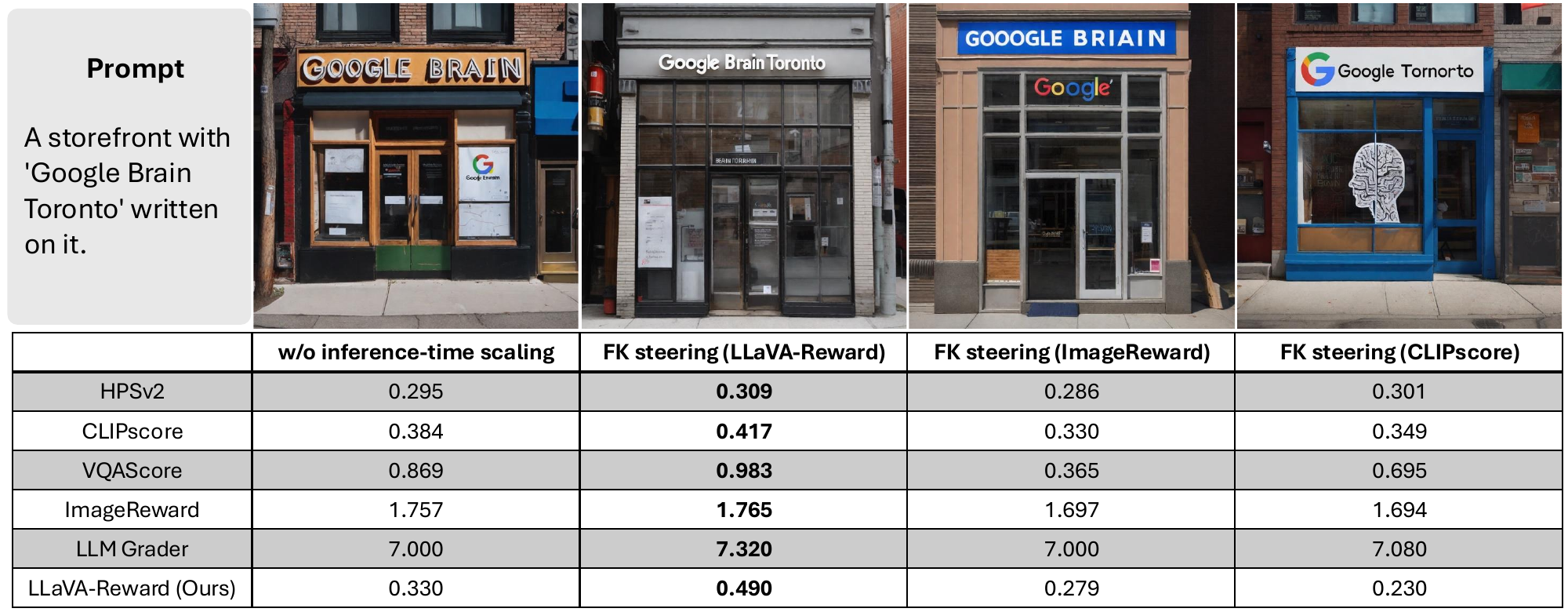}
    \includegraphics[width=0.95\linewidth]{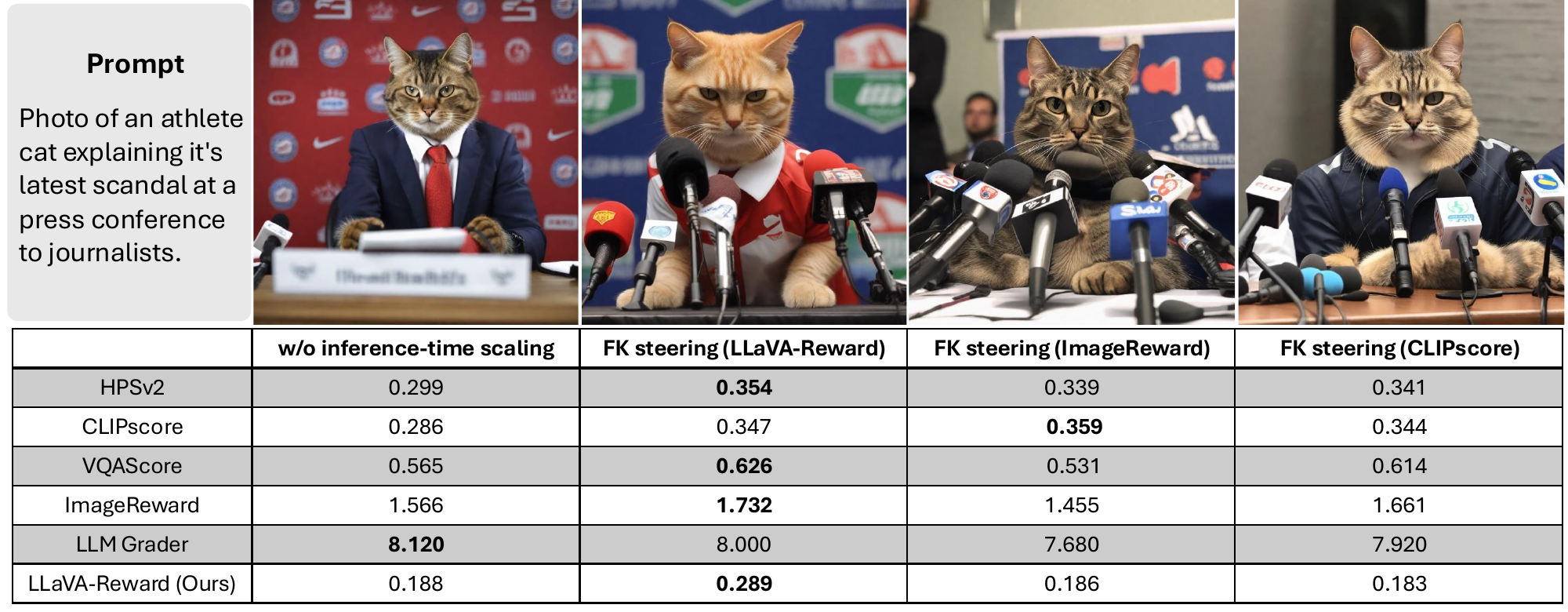}
    \includegraphics[width=0.95\linewidth]{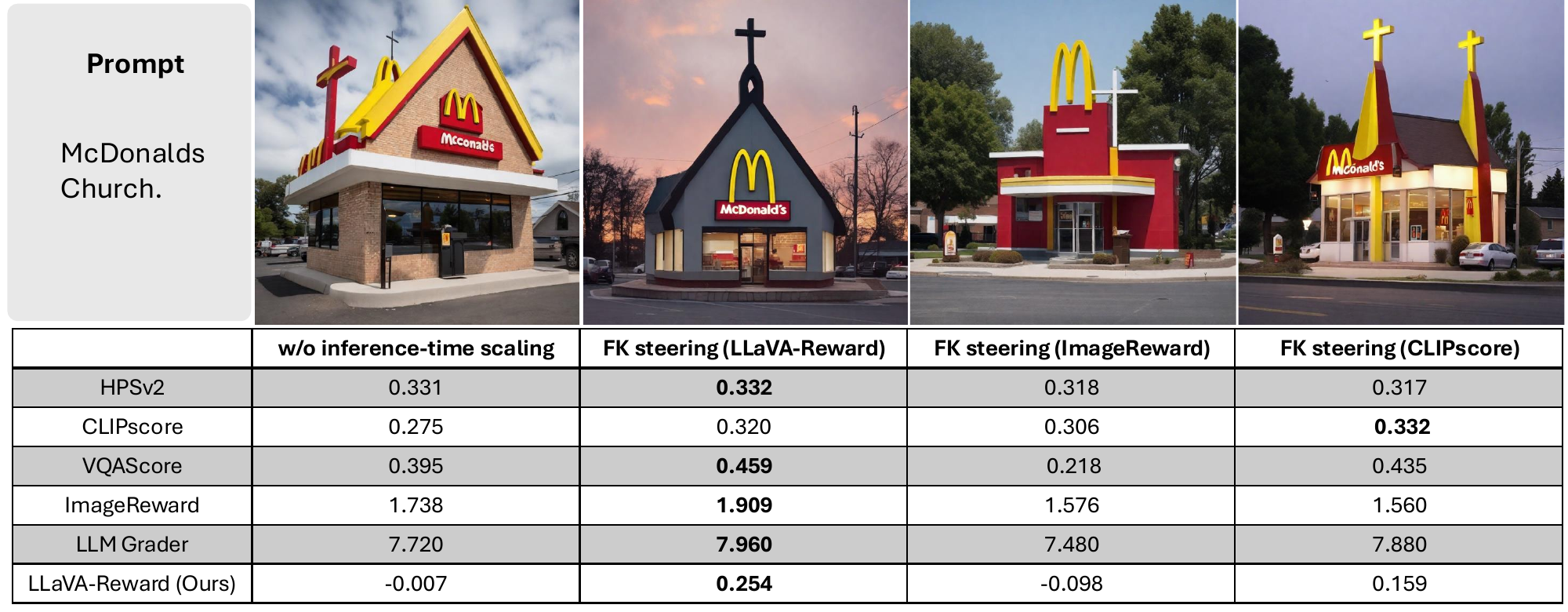}
    \vspace{-1em}
    \caption{Examples of diffusion inference-time scaling via FK steering (SDXL) using 5 different reward models with the prompt from DrawBench. The LLM grader is conducted using GPT-4o with prompts from \citet{ma2025inference}.}
    \label{fig:sample2}
    \vspace{-1.5em}
\end{figure*}

Besides the GenEval scores in \cref{tab:geneval}, we show the overall performance of 3 reward models in GenEval and DrawBench using SD v2.1 and SDXL here in \cref{tab:generation}. Additional visual examples of diffusion inference-time scaling via FK steering (SDXL) using \method{} and baselines are shown here in \cref{fig:sample0}, \cref{fig:sample1} and \cref{fig:sample2}. 

